\documentclass[letterpaper, 10 pt, conference]{ieeeconf}  

\IEEEoverridecommandlockouts                              

\usepackage{amsmath} 
\usepackage{amssymb}  
\usepackage{array}
\usepackage{booktabs}
\usepackage{graphicx}
\usepackage[section]{placeins}
\usepackage{caption}
\usepackage{multirow}
\usepackage{amsfonts}
\usepackage{subfig}
\usepackage{float}
\usepackage{xcolor}
\usepackage[bottom]{footmisc}
\usepackage{hyperref}
\makeatletter
\newcommand*\titleheader[1]{\gdef\@titleheader{#1}}
\AtBeginDocument{%
  \let\st@red@title\@title
  \def\@title{%
    \bgroup\normalfont\large\centering\@titleheader\par\egroup
    \vskip1.5em\st@red@title}
}
\makeatother

\title{\LARGE \bf
Visual-Inertial and Leg Odometry Fusion for Dynamic Locomotion}
\titleheader{
\vspace{-1.5cm}
{\small
\color{gray}
Submitted to IEEE International Conference on Robotics and Automation (ICRA), 2023}
}

\author{Victor Dh\'{e}din$^{1}$, Haolong Li$^{1}$, Shahram Khorshidi$^{2}$, Lukas Mack$^{1}$, Adithya Kumar Chinnakkonda Ravi$^{1,2}$,\\ Avadesh Meduri$^{3}$, Paarth Shah$^{4}$, Felix Grimminger$^{5}$, Ludovic Righetti$^{2,3}$, Majid Khadiv$^{2}$, and Joerg Stueckler$^{1}$
\thanks{*This work has been supported by the Max Planck Institute for Intelligent Systems through Grassroots project GR1030}
\thanks{$^{1}$ Embodied Vision Group, Max Planck Institute for Intelligent Systems,
        Tuebingen, Germany
        {\tt\small firstname.lastname@tue.mpg.de}}%
\thanks{$^{2}$ Movement Generation and Control Group, Max Planck Institute for Intelligent Systems, Tuebingen, Germany
        {\tt\small firstname.lastname@tue.mpg.de}}%
\thanks{$^{3}$ Tandon School of Engineering, New York University, New York, USA
        {\tt\small firstname.lastname@nyu.edu}}%
\thanks{$^{4}$ Oxford Robotics Institute, University of Oxford, England
        {\tt\small firstname@oxfordrobotics.institute}}%
\thanks{$^{5}$ Autonomous Motion Department, Max Planck Institute for Intelligent Systems, Tuebingen, Germany
        {\tt\small firstname.lastname@tue.mpg.de}}%
}

\begin{document}
\makeatletter
\g@addto@macro\@maketitle{
\begin{center}
    \centering
    \captionsetup{type=figure}
	\includegraphics[width=0.99\linewidth]{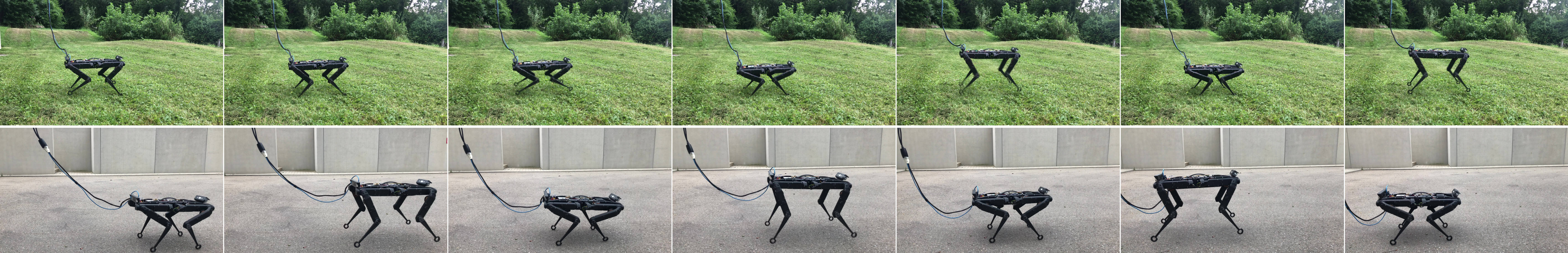}
	\captionof{figure}{Solo12 in outdoor experiments. Top: trotting and jumping on grass. Bottom: jumping from right to left on asphalt.}
	\label{fig:outdoor_results}
\end{center}
\vspace{-2ex}
}
\makeatother
\maketitle

\setcounter{figure}{1} %

\thispagestyle{empty}
\pagestyle{empty}

\begin{abstract}
Implementing dynamic locomotion behaviors on legged robots requires a high-quality state estimation module. 
Especially when the motion includes flight phases, state-of-the-art approaches fail to produce reliable estimation of the robot posture, in particular base height. 
In this paper, we propose a novel approach for combining visual-inertial odometry (VIO) with leg odometry in an extended Kalman filter (EKF) based state estimator. 
The VIO module uses a stereo camera and IMU to yield low-drift 3D position and yaw orientation and drift-free pitch and roll orientation of the robot base link in the inertial frame. 
However, these values have a considerable amount of latency due to image processing and optimization, while the rate of update is quite low which is not suitable for low-level control. 
To reduce the latency, we predict the VIO state estimate at the rate of the IMU measurements of the VIO sensor.
The EKF module uses the base pose and linear velocity predicted by VIO, fuses them further with a second high-rate IMU and leg odometry measurements, and produces robot state estimates with a high frequency and small latency suitable for control. 
We integrate this lightweight estimation framework with a nonlinear model predictive controller and show successful implementation of a set of agile locomotion behaviors, including trotting and jumping at varying horizontal speeds, on a torque-controlled quadruped robot.

\end{abstract}

\section{INTRODUCTION}
Legged robots are potentially capable of traversing uneven and unstructured terrains through making and breaking contacts with their environments using their feet and hands. However, this capability introduces new challenges for estimation and control algorithms. For instance, an estimation algorithm should constantly fuse the exteroceptive and proprioceptive measurements with the kinematics of the limbs currently in contact with the environment to estimate the robot floating base posture and velocity for low-level control. 

Early works for base state estimation of legged robots focused on fusing an on-board IMU with the leg odometry through an extended Kalman filter (EKF) framework to provide estimates of base states for the low-level controller \cite{bloesch2012_kinimufuse,bloesch2013_ekflegvel,rotella2014state}. While this approach can provide drift-free base velocity and roll-pitch orientation, the base position and yaw orientation are unobservable which poses limitations especially for locomotion on uneven surfaces or motions with considerable vertical motion of the base (such as jumping, see Fig.~\ref{fig:outdoor_results}).

Recent works couple these proprioceptive measurements with exteroceptive modalities, e.g., camera or Lidar, through loosely \cite{camurri2020_pronto} or tightly \cite{hartley2018hybrid,wisth2021_vilens} coupled methods. While the tightly coupled approach has the benefit of fusing all the modalities with direct consideration of their measurement uncertainty, it can be computationally very demanding especially for robots with limited compute budget.
In our approach, we aim at a loosely coupled approach to integrate visual-inertial state estimation with leg odometry in a high-rate EKF state estimator to provide low-drift states which are sufficiently accurate and smooth for control. 
This way, the EKF and controller computation can run on a different device than the visual-inertial odometry (VIO).
Furthermore, we can predict the VIO measurements and use them to reduce the delay, while the EKF can access the low-drift pose estimates from VIO. 
The main contributions of this work are 
    1) We propose a novel approach to combine the benefits of VIO and leg odometry in a loosely coupled EKF approach to estimate low-latency and low-drift base states for agile locomotion. We compensate for height drift of the VIO using leg kinematics measurements when the legs are in contact with the ground.
    2) We perform an extensive set of experiments including outdoors on the open-source quadruped Solo12 \cite{grimminger2020open}. This is the first work that integrates visual and proprioceptive measurements with nonlinear model predictive control for dynamic locomotion on this hardware.

\section{RELATED WORK}

State estimation from only leg odometry and IMU such as in~\cite{bloesch2012_kinimufuse,hartley2020_contactaidekf,hartley2018_preintcont,kim2021_dyncontact} has limitations in observability of state variables such as yaw rotation or absolute position in a world reference frame.
To this end, several approaches combine proprioceptive and IMU measurements with exteroceptive sensors such as vision~\cite{chilian2011_multisensorfusion,hartley2018_hybridvic,dudzik2020_robnavquad,teng2021_legslipinekf,kim2022_step}, LiDAR~\cite{nobili2017_sensfuse}, or both~\cite{camurri2020_pronto,wisth2019_robustestfact}.
Vision sensors are particularly lightweight compared to LiDARs. 
They typically impose only little constraints on the payload of the quadruped which is particularly important for dynamic quadrupeds.
Chilian et al.~\cite{chilian2011_multisensorfusion} proposed an early multi-sensor fusion approach which integrates IMU pose measurements with relative pose measurements from visual and leg odometry. 
The pose information is combined in a weighted manner.
Teng et al.~\cite{teng2021_legslipinekf} extend an EKF approach which fuses IMU and leg odometry to also integrate velocity measurements from a visual-inertial odometry method.
In~\cite{hartley2018_hybridvic} a fixed-lag smoothing approach based on factor graph optimization has been proposed. The approach uses visual odometry estimates as relative pose factors.
Kim et al.~\cite{kim2022_step} tightly integrate visual keypoint depth estimation with inertial measurement and preintegrated leg velocity factors.
Our approach integrates absolute yaw and position measurements by the VIO, while height drift of the VIO wrt. the ground height is compensated by estimating the height bias in the EKF.
In our approach, we aim at a lightweight system which decouples visual-inertial state estimation from the high-rate EKF state estimator used for control. 
This way, EKF and controller can run on a different compute device than the VIO.
Moreover, by predicting the VIO measurements, delay is reduced and computational load for reintegration of measurements in the EKF can be avoided.

\section{METHOD}

\begin{figure}
	\centering
	\includegraphics[width=0.99\linewidth]{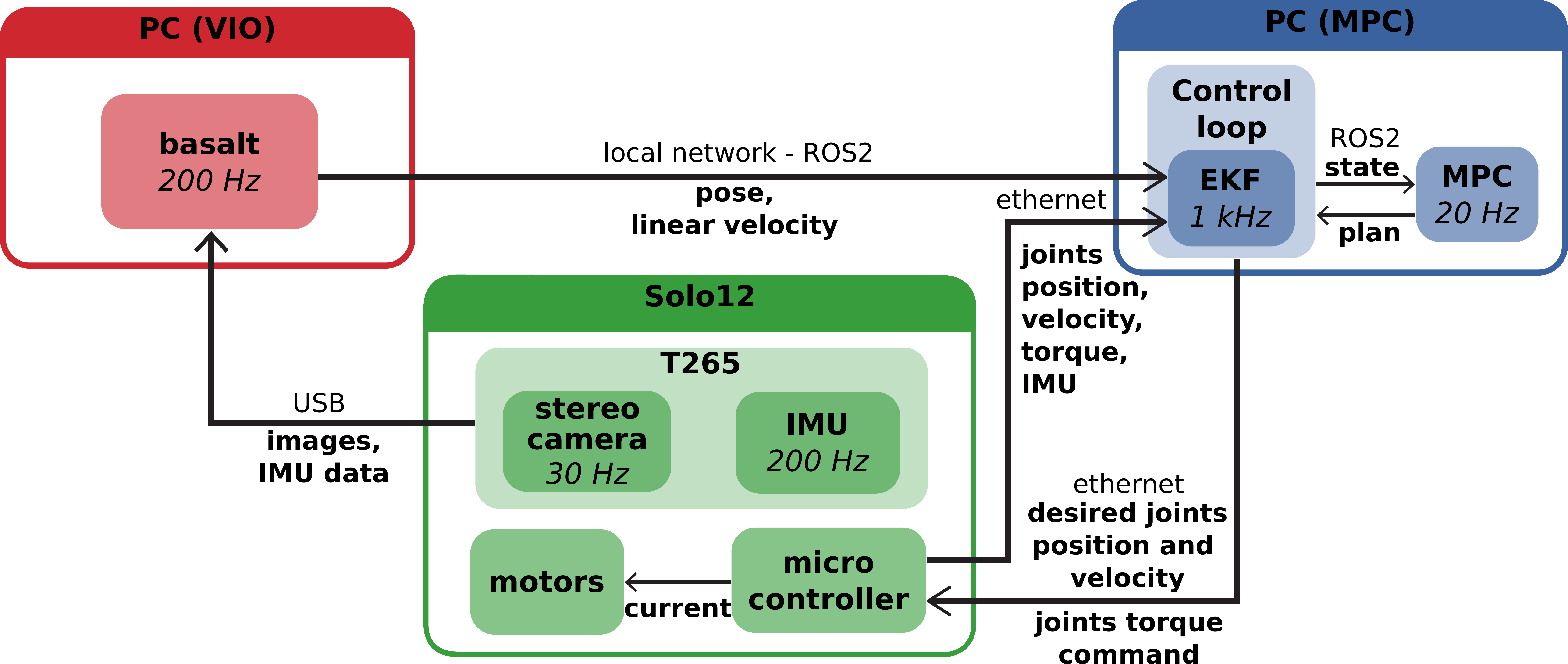}
	\caption{System overview and communication diagram.}
	\label{fig:communication_diagram}
\end{figure}

\begin{figure}
	\centering
	\subfloat[Robot base and camera frames.]{\includegraphics[width=.45\linewidth]{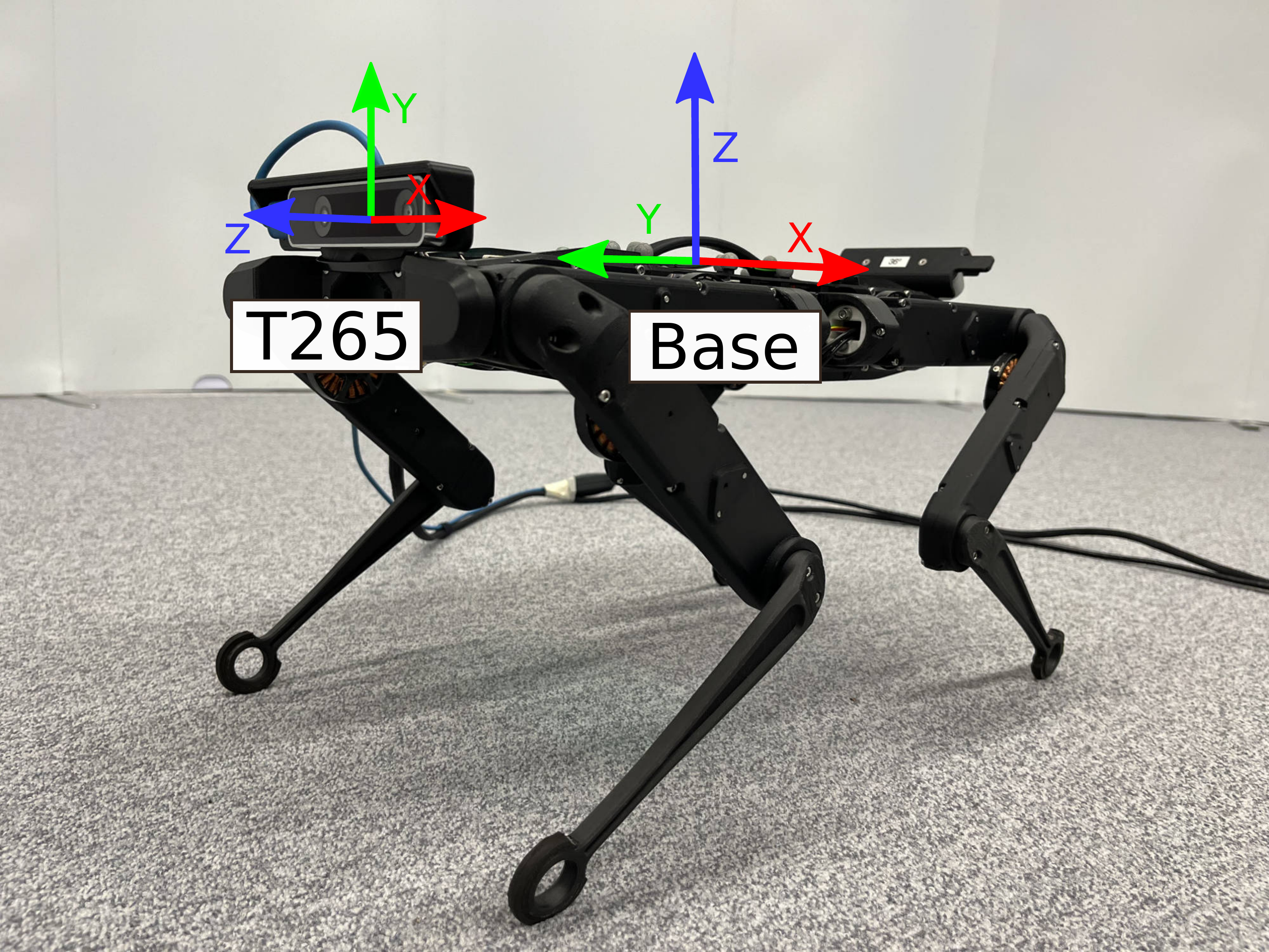}}  \label{fig:1a}
	\subfloat[Graphical model.]{\includegraphics[width=.53\linewidth]{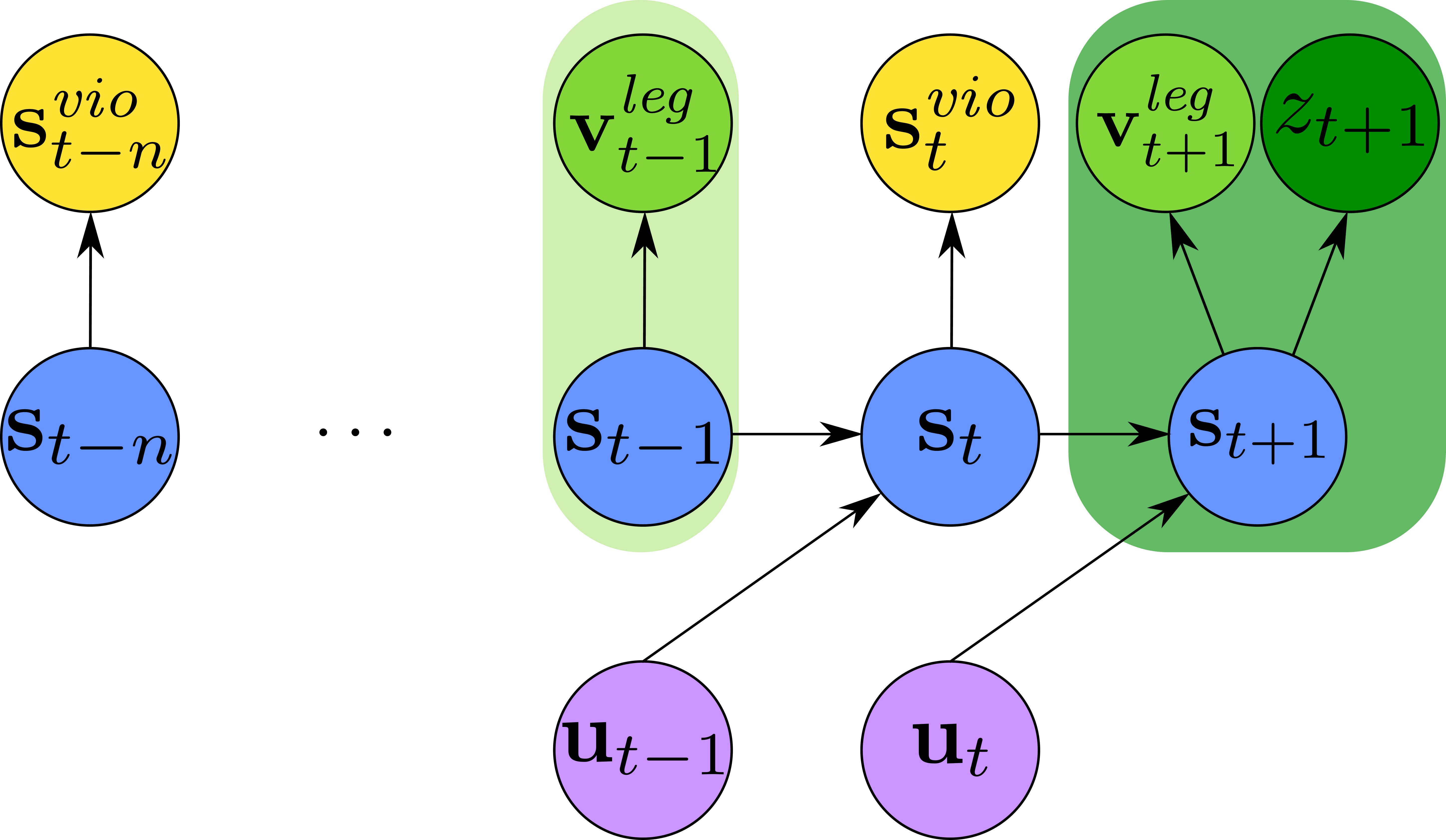}}  \label{fig:1b}
	\caption{Left: robot base and camera frames. Right: graphical model. The x axis of robot base points forwards and z axis points upwards. The state of the EKF is represented as $\mathbf{s}_t$ in blue circle. We use the measurement of IMU (circle in magenta) mounted on the robot to predict next state at 1000\,Hz. The yellow circle represents VIO measurement at 200\,Hz, the shallow green circle is the leg velocity measurement at contact. The height measurement in dark green circle is added if all four legs are in contact with ground.}
	\label{fig:graph}
	\vspace{-0.3cm}
\end{figure}

In our approach, we fuse visual and inertial measurements with leg odometry for estimating the position, orientation, and velocity of the robot with respect to the ground plane. Figure~\ref{fig:communication_diagram} provides an overview of our system. 
Base state estimation is performed at high frequency with low latency using an EKF to be used in a real-time model-predictive control (MPC) approach for trotting and jumping motions~\cite{meduri2022_biconmp}. The EKF fuses information from different sensory sources (see Fig.~\ref{fig:graph}b): it takes as input measurements of an Inertial Measurement Unit (IMU) mounted on the robot, leg odometry data from the joints of the legs (angular position, angular velocity and torque), and pose measurements estimated by a visual-inertial odometry (VIO) algorithm (using a second IMU in a visual-inertial camera).

For dynamic locomotion, accurately estimating the height of the robot above the ground plane is important. 
While the VIO does not provide an absolute reference to the ground plane directly and will drift in height over time, we use contact detection and leg kinematics to obtain height measurements. In fact, VIO and leg odometry provide complementary strengths. VIO can measure the absolute roll and pitch in the environment, and build a map of the environment for estimating the base position and yaw orientation (rotation around gravity direction) with respect to this local map.
While the local map estimate still drifts, this estimation error is typically significantly smaller than those obtained by the leg odometry which is prone to foot slippage and inaccuracies of the contact detection.


\subsection{Visual-Inertial Odometry}

VIO algorithms estimate the motion of a camera over time by tracking landmarks detected in the successive camera images from one or several cameras and integrating inertial measurement from an IMU using kinematics. 
This problem is usually formulated as finding a state that minimizes both a reprojection error term $E_{V}(\mathbf{s})$ computed on landmarks and an error term $E_{I}(\mathbf{s})$ associated with the motion determined from the IMU measurements,
\begin{equation}
\label{eq:viominprob}
   \mathbf{s}_\mathit{VIO}^{*} = \operatorname{arg\,min}_{\mathbf{s}} E_{V}(\mathbf{s}) + E_{I}(\mathbf{s}).
\end{equation}
We base our VIO estimator on basalt~\cite{usenko2020_basalt}.
It estimates the VIO sensor state  
    $\mathbf{s}_\mathit{VIO} = \left( {^W}\mathbf{p}_{WB}, {^W}\mathbf{q}_{WB}, {^W}\mathbf{v}{_{WB}}, \mathbf{b}_i^a, \mathbf{b}_i^g, \mathbf{l}_1, \ldots, \mathbf{l}_N \right)$,
where ${^W}\mathbf{p}_{WB} \in \mathbb{R}^3$ is the robot base link position expressed in the VIO world frame $W$, ${^W}\mathbf{q}_{WB} \in S^3$ is the robot base link orientation in world frame, ${^W}\mathbf{v}{_{WB}} \in \mathbb{R}^3$ is the linear velocity of the robot with respect to world expressed in world frame. $\mathbf{b}_i^a, \mathbf{b}_i^g \in \mathbb{R}^3$ are the accelerometer and gyroscope biases, respectively.
The landmarks $\mathbf{l}_i$ are 3D coordinates of detected and matched keypoints parametrized by 2D image plane coordinates and the inverse distance $d_l$ in the hosting image frame.

The reprojection error term is defined as a weighted sum of squared residuals over a set of keypoint observations of the landmarks in multiple frames. 
A KLT tracking method~\cite{shi1994_klt} is used to detect and track a sparse set of keypoints between frames. 
If the proportion of new keypoints is above a threshold, the frame becomes a keyframe.
%
The IMU error term is computed by comparing a pseudo measurement $\Delta s = (\Delta \mathbf{R}, \Delta \mathbf{v}, \Delta \mathbf{p})$, which corresponds to several consecutive IMU measurements integrated between two frames $i$ and $j$ of respective timestamps $t_i$ and $t_j$, to the pose of the state at time $t_{i}$ and $t_{j}$.
For each IMU data at time $t$ ($t_i < t \le t_j$) the precedent measurement is updated using the bias corrected IMU acceleration $ \mathbf{a}_t = \mathbf{a}_t^{IMU} - \mathbf{b}_i^a $ and the bias corrected IMU angular velocities $ \mathbf{\omega_t} = \mathbf{\omega}_t^{IMU} - \mathbf{b}_i^g $ as follows
\begin{equation}\label{eq:preintegration}
\begin{split}
\Delta \mathbf{R}_{t+1} &= \Delta \mathbf{R}_{t} \exp ( \mathbf{\omega}_{t+1} \Delta t)\\
\Delta \mathbf{v}_{t+1} &= \Delta \mathbf{v}_{t} + \Delta \mathbf{R}_{t} \mathbf{a}_{t+1} \Delta t\\ 
\Delta \mathbf{p}_{t+1} &= \Delta \mathbf{p}_{t} + \mathbf{v}_{t} \Delta t. 
\end{split}
\end{equation}
The residuals are the difference between preintegrated relative pose measurement and the relative pose between two consecutive frames~\cite{usenko2020_basalt}. 

\paragraph{Windowed Optimization} 
The reprojection error (left) in Eq.~\eqref{eq:viominprob} is computed over a set of keypoints that are observed in different frames. 
To prevent the size of the optimization problem from growing, basalt uses a bounded window of recent frames and keyframes, and marginalizes information of old frames that drop out of the optimization window. 
In our case, the window corresponds to the 3 most recent frames and 7 most recent keyframes. 

\paragraph{Low-Latency VIO Prediction} In practice, the VIO has a moderate latency due to computation (approximately 5.8 ms optimization time on average with a standard deviation of 3.1 ms in our setup) and additional communication delays. 
The output rate is limited by the image frame rate.
We propose to use IMU predictions to update the last VIO state estimate at a higher rate and to fuse these output states with leg odometry and a high precision IMU on the robot which helps reducing the latency and increasing the output rate.
By this, also computation time can be saved for the EKF which would otherwise require memorization of old states and measurements, and reestimation after each image-rate update on the EKF/MPC compute device (as e.g. in~\cite{camurri2020_pronto}).
The VIO outputs the prediction of the robot pose and velocity at the rate of the IMU in the VIO sensor (200\,Hz for our camera) estimated using the IMU preintegration model in Eq.~\eqref{eq:preintegration} from the latest camera frame with optimization result available.
Once the optimization result for the current frame is available, we reintegrate the IMU measurements and continue predicting the VIO state from this updated pose estimate (on the VIO compute device).

\subsection{Sensor Fusion for Legged Robot State Estimation}
We adapt the approach in~\cite{camurri2020_pronto} to fuse measurements of the pose and velocity of the robot's base link using an Extended Kalman Filter (EKF).
Differently to~\cite{camurri2020_pronto} we integrate high-rate, low latency state observations from VIO and estimate the difference between VIO height estimate and ground height by leg kinematics.
The EKF allows for integrating measurements with various rates and asynchronous timing. 
The state estimated by the EKF is
$\mathbf{s}_{\mathit{EKF}} = \left( {^W}\mathbf{p}_{WB}, {^W}\mathbf{q}_{WB}, {^B}\mathbf{v}{_{WB}}, \mathbf{b}_i^a, \mathbf{b}_i^\omega, \mathbf{b}^{\delta z}\right)$,
where ${^W}\mathbf{p}_{WB} \in \mathbb{R}^3$, ${^W}\mathbf{q}_{WB} \in S^3$, ${^B}\mathbf{v}{_{WB}} \in \mathbb{R}^3$ are position, orientation, and linear velocity of the robot's base link in the world frame, and $\mathbf{b}_i^a$ and $\mathbf{b}_i^\omega$ are the biases of IMU accelerometer and gyroscope measurements from an IMU mounted on the robot base (different IMU than used for VIO).
The height bias $\mathbf{b}^{\delta z}$ compensates for the vertical drift of the VIO.
We use the IMU prediction model in~\cite{camurri2020_pronto} to propagate the state with the IMU measurements and estimate the acceleration and gyro biases.

\subsubsection{Leg Odometry Measurements}
By determining the set of feet in contact with the ground, we can measure the linear velocity of the robot's base link from the leg kinematics. By assuming that the foot $k$ remains stationary while it is in contact with the ground, the linear velocity of the floating base can be measured as~\cite{camurri2020_pronto}
\begin{equation} \label{equ:ekf_measurement}
    {^B}\mathbf{v}_{WB} = -{^B}\mathbf{v}_{BK} - {^B}\boldsymbol{\omega}_{WB} \times {^B}\mathbf{p}_{BK}
\end{equation}
This method enables a good accuracy on velocities and low latency. However, since only the velocity is observable, this method hardly handles drift in position, especially in height, which is detrimental for control, especially for motions with significant changes in base height.
The angular velocity in this observation model is measured directly by the IMU compensated with the estimated gyroscope bias. 

We choose a simpler contact classification model than~\cite{camurri2020_pronto} in order to estimate the set of feet in contact. 
By assuming that the robot base remains flat during contact transitions, we can consider an equal distribution of the robot's total weight over the feet in contact with the ground. 
We use a Schmitt trigger to implement a robust hysteresis on the contact decision.
If the norm of the force at each endeffector is higher than an upper threshold, we consider the foot as in contact with the ground, and if the norm is below a lower threshold, the endeffector is no longer in contact. 
The hysteresis in the contact detection helps to reject outliers due to high joint acceleration when the endeffector leaves the ground. 
We compute the endeffector force norm $F_K = ||\mathbf{F}_K||$ using the joint torque by
    $\mathbf{F}_K = (\mathbf{S}_K \mathbf{J}_K^T)^{-1} \mathbf{S}_K \boldsymbol{\tau}$,
where $\mathbf{S}_K$ is the selection matrix for the joints of leg $k$.
To further exclude outliers, the leg odometry measurement is updated only if the leg is in contact with the ground for $N_{contact}$ consecutive steps. 

By having the joint positions and velocities sensed from the encoders one can use forward kinematics to compute the velocity and position of each endeffector in the base frame. By collecting all the effects of noise into one additive noise term, the measurement model can be rewritten as~\cite{camurri2020_pronto}
$-\boldsymbol{J}(\Tilde{\mathbf{q}}_k)\Tilde{\dot{\mathbf{q}}}_k - \boldsymbol{\omega} \times \operatorname{fk}(\Tilde{\mathbf{q}}_k) = {^{B}}\mathbf{v}{_{WB}^{\mathit{EKF}}} + \boldsymbol{\eta}^v$
where $\Tilde{\mathbf{q}}_k$ are the measured joint angles of leg~$k$, and $\operatorname{fk}(\Tilde{\mathbf{q}}_k) = {^B}\mathbf{p}_{BK}$ is the forward kinematics for the foot contact point.

\subsubsection{VIO Pose and Velocity Measurements}
The VIO provides additional pose and velocity estimates of the robot base link in the inertial frame (world frame).
Roll and pitch are estimated drift-free by the VIO, while 3D position and yaw orientation are estimated with respect to the estimated keypoint map and can drift.
However, the drift in position and yaw orientation is significantly smaller than the drift by fusing leg odometry and IMU alone.
The measurement model of the VIO pose and velocity is
\begin{equation}
\begin{split}
    {^{W}}\mathbf{p}{_{WB}^{\mathit{VIO}}} &= {^{W}}\mathbf{p}{_{WB}^{\mathit{EKF}}} + ( 0, 0, \mathbf{b}_{\delta z} )^\top + \boldsymbol{\eta}^{p}\\
    {^{W}}\mathbf{\theta}{_{WB}^{\mathit{VIO}}} &= {^{W}}\mathbf{\theta}{_{WB}^{\mathit{EKF}}} + \boldsymbol{\eta}^{\theta}\\
    {^{W}}\mathbf{v}{_{WB}^{\mathit{VIO}}} &= \mathbf{R}\!\left({^{W}}\mathbf{q}{_{WB}}\right) {^{B}}\mathbf{v}{_{WB}^{\mathit{EKF}}} + \boldsymbol{\eta}^{v},
\end{split}
\end{equation}
where ${^{W}}\mathbf{\theta}{_{WB}}$ is the orientation of the base in world frame expressed in $\mathfrak{so}(3)$.
To tackle drift of the VIO in the height estimate, we estimate a height bias $\mathbf{b}_{\delta z}$ which is the difference of the measured height of the base link above the ground and the estimated height by the VIO.


\subsubsection{Ground Height Measurements}
The ground height is only measured when all the legs are considered as being in contact with the ground. 
The ground height is measured as the average of the height measurements by the different legs which is computed by forward kinematics, i.e.
\begin{equation}
\begin{split}
    {^W}z_{WB}(\Tilde{\mathbf{q}}) &:= 
    \left[ \frac{1}{N_{legs}} \sum_{i=1}^{N_{legs}} -\operatorname{fk}(\Tilde{\mathbf{q}}_i) \right]_2,
\end{split}
\end{equation}
where $N_{legs}$ is the number of legs in contact ($N_{legs} = 4$ in our case), 
and the operator $\left[ \cdot \right]_k$ selects the $(k+1)$-st entry of a vector.
Additionally, to exclude outliers and inaccurate measurements, the ground height is measured only after all the legs are considered as in contact with the ground for $N_{\mathit{standing}}$ consecutive steps. 
The measurement model for the EKF is
    $\Delta z  = \mathbf{b}_{\delta z} + \boldsymbol{\eta}^{\mathbf{b}_{\delta z}}$
with additive Gaussian noise $\boldsymbol{\eta}^{\mathbf{b}_{\delta z}}$.
The height bias is measured through the joint angle and the VIO pose measurements by $\Delta z = \left[ {^{W}}\mathbf{p}{_{WB}^{\mathit{VIO}}}\right]_2 -{^W}z_{WB}(\Tilde{\mathbf{q}})$.

\subsection{Control Architecture}
We use the non-linear MPC developed in~\cite{meduri2022_biconmp} to control the robot. 
The MPC requires a contact plan as input and determines whole-body trajectories for the robot. In this work, we only consider cyclic gaits, e.g., trotting and jumping, where the contact plan is automatically generated based on a command linear velocity (sidewards and forwards/backwards motion at a constant yaw angle). In this case, the Raibert heuristics is used to adapt the contact locations based on the feedback of the base linear velocity~\cite{meduri2022_biconmp}. The framework generates centroidal trajectories using alternating direction method of multipliers (ADMM) approach and then a differential dynamic programming (DDP) based kinematic optimizer is used to generate desired joint trajectories. Using an unconstrained inverse dynamics, the desired joint torques are computed and fed to the robot joint controller at 1\,kHz. 

\section{EXPERIMENTS}


\begin{figure*}[tb]
    \centering
    \includegraphics[width=0.84\linewidth]{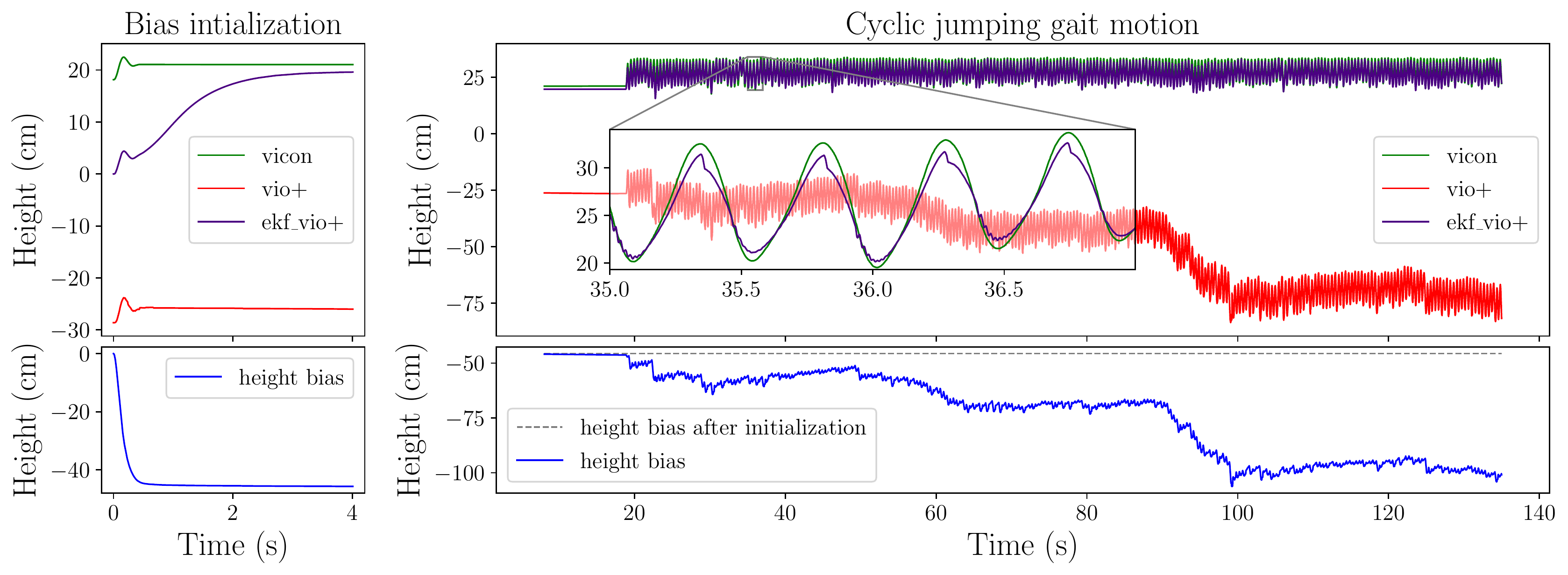}
    \caption{Height estimate of VIO with IMU predictions (vio+) and our approach (ekf\_vio+) compared with ground truth for jumping. Left: initialization (standing), right: jumping. The fast decline in the flight phase is due to false contact detection.}
    \label{fig:jump_height}
\end{figure*}

\begin{figure}
	\centering
	\includegraphics[width=0.99\linewidth]{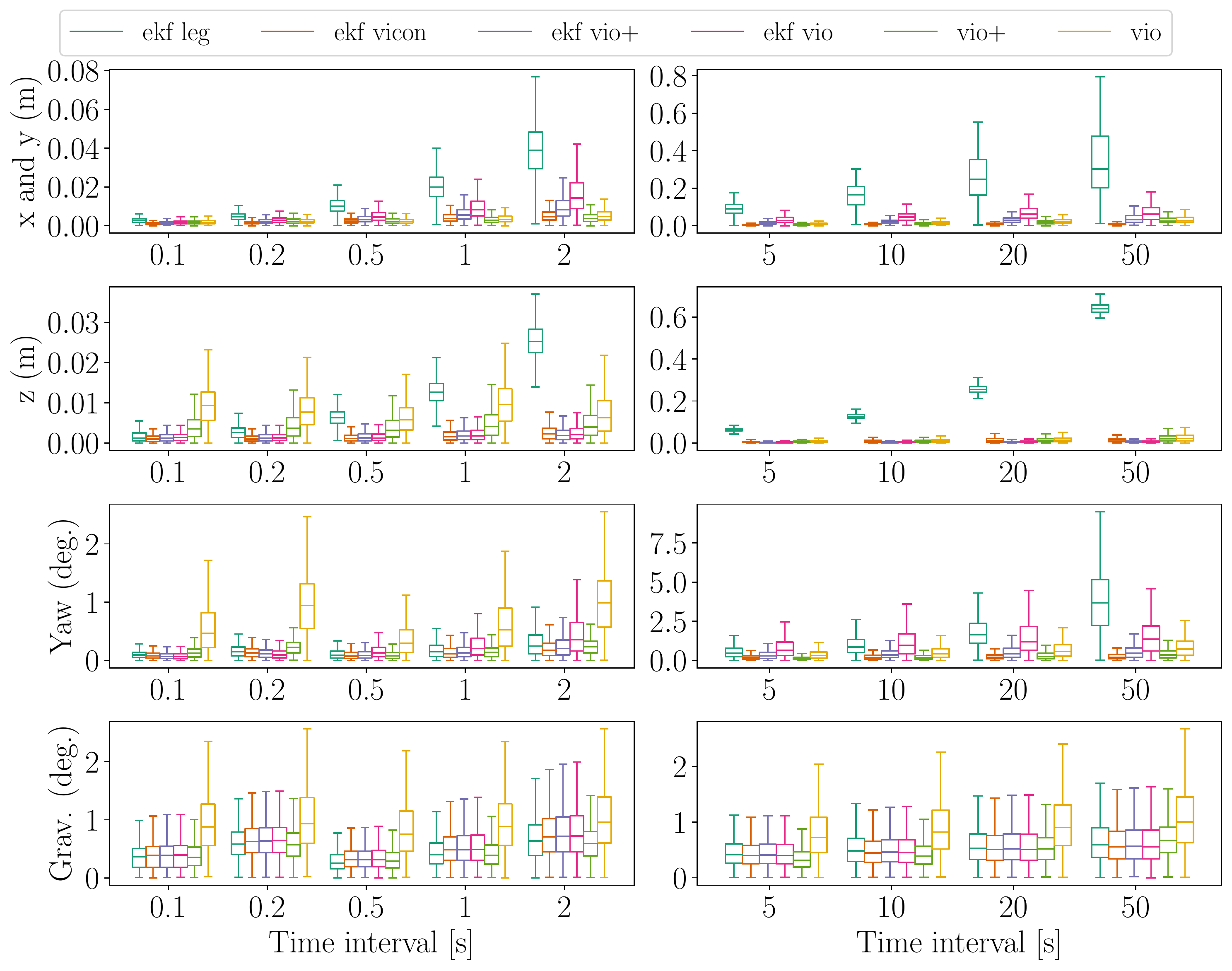}
	\caption{Trotting RPE for all time intervals.}
	\label{fig:rpe_trotting}
\end{figure}

\begin{table}
	\caption{Trotting trajectory accuracy in RPE. 
	}
	\label{tab:result_trotting}
	\centering
		\resizebox{\linewidth}{!}{
		\begin{tabular}{cccccccc}
			\toprule
			& & ekf\_leg & ekf\_vicon & ekf\_vio+ & ekf\_vio  & vio+ & vio\\
			\midrule
            \multirow{2}{*}{x and y\,(m)}& mean & 0.151 & 0.011 & 0.022 & 0.041 & 0.016 & 0.018\\
            &max & 0.794 & 0.264 & 0.155 & 0.237 & 0.138 & 0.143 \\
			\midrule
            \multirow{2}{*}{z\,(m)}& mean & 0.198 & 0.009 & 0.005 & 0.005 & 0.011 & 0.014\\
            &max & 0.746 & 0.098 & 0.057 & 0.058 & 0.076 & 0.107 \\
			\midrule
			\multirow{2}{*}{yaw\,(deg)}& mean & 1.440 & 0.466 & 0.400 & 1.044 & 0.269 & 0.816 \\
			& max & 9.532 & 6.707 & 2.563 & 7.588 & 1.670 & 3.086 \\
			\midrule
			\multirow{2}{*}{gravity\,(deg)}& mean &0.593 & 0.615 & 0.632 & 0.637 & 0.545 & 1.117 \\
			& max & 2.256 & 2.593 & 2.783 & 2.802 & 2.109 & 4.878 \\
			\bottomrule
		\end{tabular}
		}
\end{table}

We evaluate our approach with the torque-controlled quadruped platform Solo12 by the Open Dynamic Robot Initiative~\cite{grimminger2020open} which we augment with a Intel Realsense T265 stereo-inertial sensor (see Fig.~\ref{fig:graph}).
Stable trotting and jumping motions are generated by the MPC~\cite{meduri2022_biconmp} which uses our state estimate and calculates joint commands.
The communication diagram is illustrated in Fig.~\ref{fig:communication_diagram}.
The robot communicates joint measurements and targets via Ethernet with the robot control PC (Intel Xeon CPU E5-1680@3.40GHz, 8 cores) which runs Linux with a real-time kernel.
A second vision PC (Intel Xeon CPU E5-1630@3.70GHz, 8 cores) computes visual-inertial odometry.
The visual-inertial odometry result is communicated to the robot control PC via Ethernet.
The Intel T265 camera provides 3-axis accelerometer and gyroscope data at 62.5\,Hz and 200\,Hz, respectively. 
The accelerometer data is upsampled to match the gyroscope measurement rate.
The sensor also provides fisheye stereo images with a wide field of view (ca. 173 degrees) at a frame rate of 30\,Hz.
We use the calibration tools of~\cite{usenko2020_basalt} to calibrate the camera intrinsic and the extrinsics of camera wrt. IMU, and the relative location of the IMU wrt. the robot base link. The orientation of the IMU wrt. the robot base link is taken from the CAD model.
For wheeled robots it has been shown that the accelerometer biases are unobservable if the robot does not move sufficiently in yaw~\cite{wu2017_vinsonwheels}.
Since the robot maintains a fixed yaw rotation, we fix the biases after a short initialization phase in which the robot is moved with 6 degrees of freedom before each run.  

We validate our approach in both indoor and outdoor environments. 
For indoor environments, we collect ground-truth data with a Vicon motion capture system at the rate of 1\,kHz. 
VIO at 30\,Hz is denoted as \emph{vio} in the following tables and figures, while VIO with IMU prediction at 200\,Hz is denoted as \emph{vio+}. 
For evaluation, both VIO versions are upsampled to the EKF rate of 1\,kHz using the latest available estimates to demonstrate the performance of using these estimates as input for the controller.
Note that our approach is not directly comparable to previous approaches such as Pronto~\cite{camurri2020_pronto}, since we propose a lightweight fusion method tailored to our control system.
Our system uses VIO predictions to avoid computations for rolling back the EKF and to leave as much compute for the controller as possible.
We use $N_{\mathit{contact}} = 1$ and $N_{\mathit{standing}} = 3$ in our experiments.  

\subsection{Evaluation Metrics}
\begin{figure}
    \centering
    \includegraphics[width=0.49\linewidth]{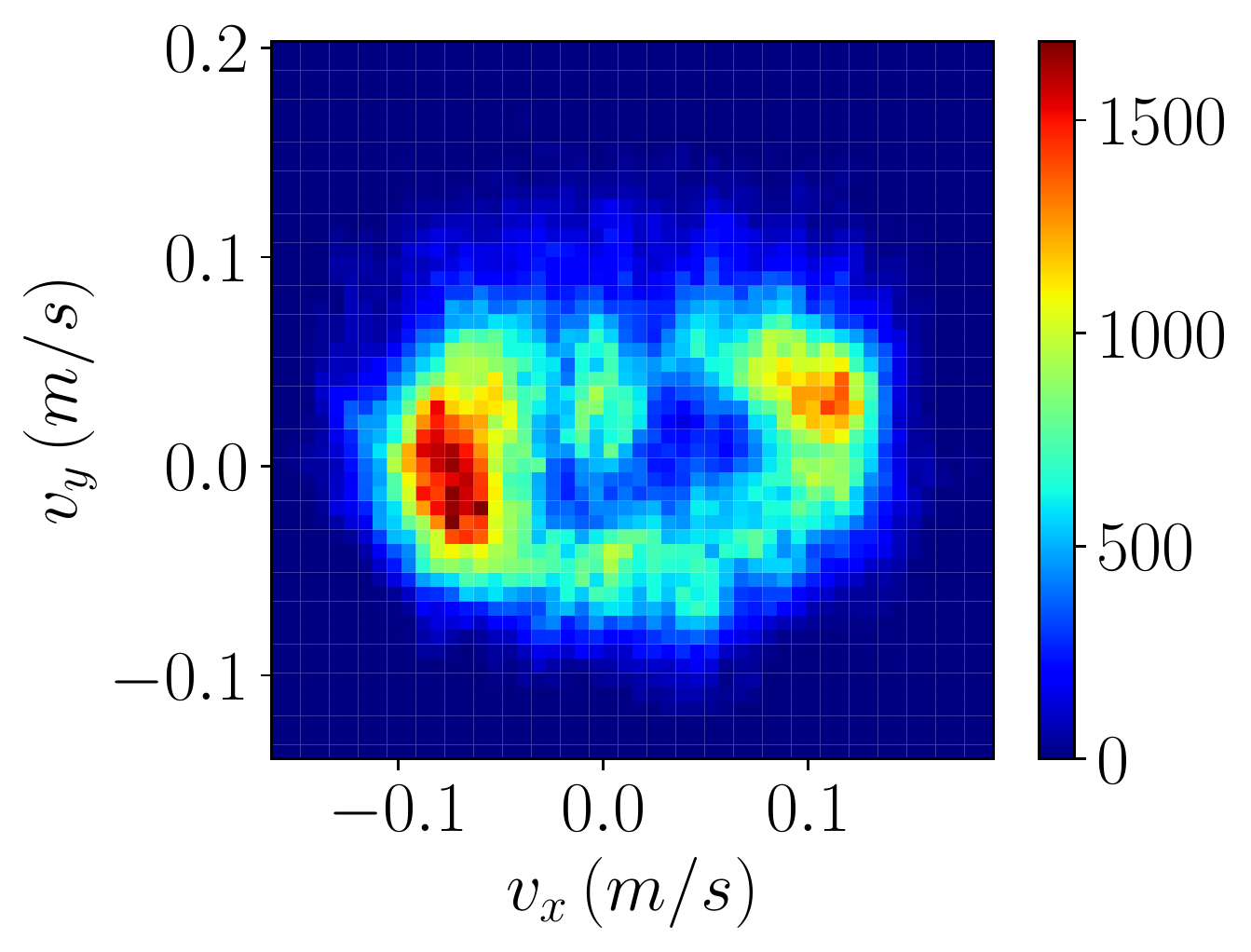}
    \includegraphics[width=0.49\linewidth]{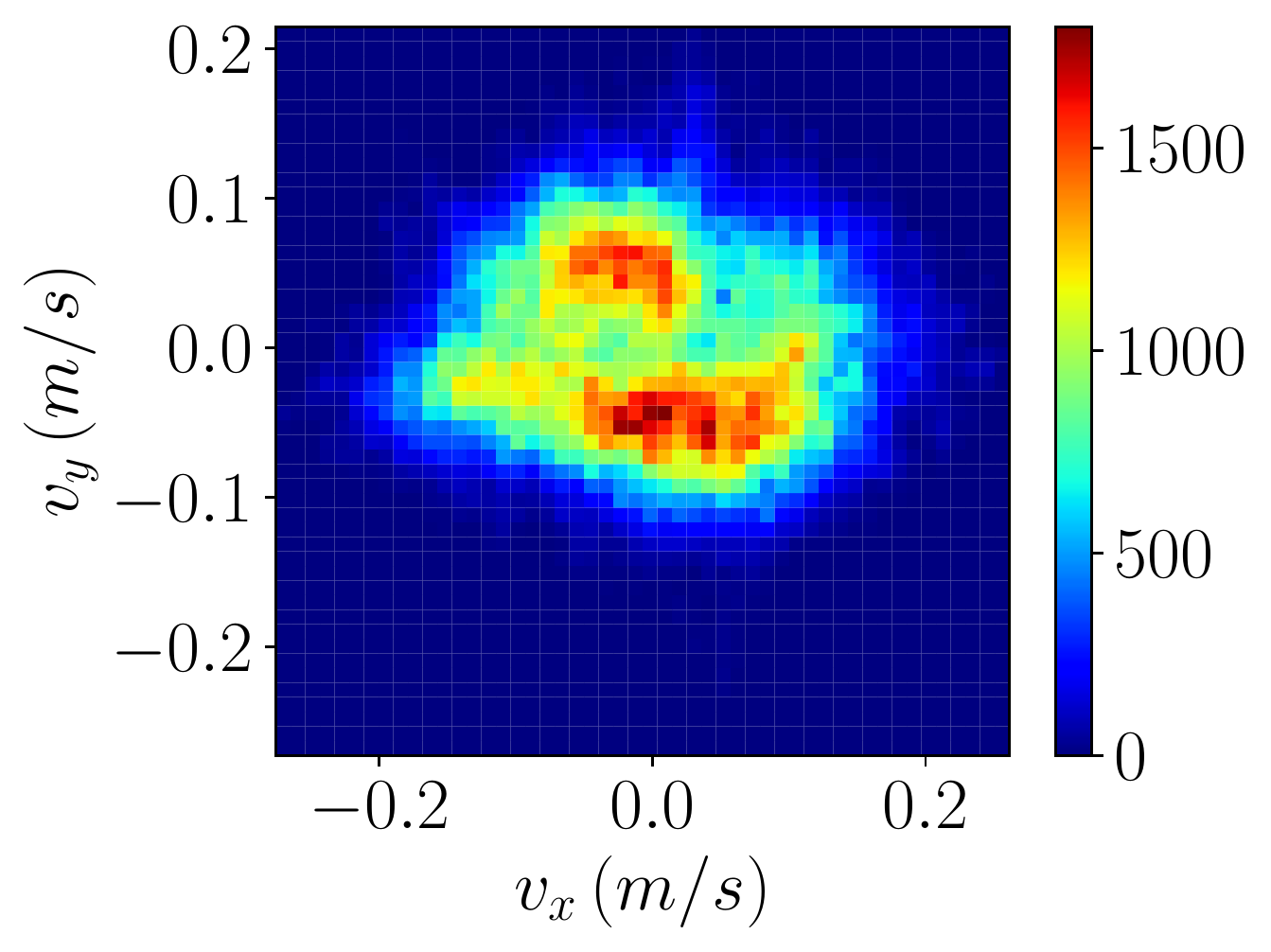}
    \caption{
Distribution of horizontal linear velocity (m/s) of the base in experiment runs (left: trot, right: jump). The velocities are determined by fusing Vicon and IMU measurements in the EKF to obtain smoothed estimates. Min./max. are at the histogram boundaries. 
The Pearson correlation coefficients between estimated and control velocities are: Trot: 0.96 in x, 0.79 in y; Jump: 0.86 in x, 0.85 in y.
According to the estimate, the robot follows the command partially due to competing MPC objectives, constraints, and Raibert heuristics for the contact plan  (Trot: factor 0.49 in x, 0.32 in y. Jump: 1.23 in x, 0.72 in y).
}
    \label{fig:vel}
\end{figure}

Since the control performance relies on the accuracy of state estimation, we evaluate the robot trajectory quantitatively using the relative pose error~(RPE~\cite{zhang2018_rpe}) metric with various subtrajectories of time intervals \{0.1, 0.2, 0.5, 1, 2, 5, 10, 20, 50\} in seconds.
We record 5 runs for each gait type (approx. 2\,min per run for trotting and jumping) at varying target horizontal linear velocity using the EKF with augmented VIO measurements for state estimation. 
Figure~\ref{fig:vel} shows the distribution of the horizontal velocity as estimated by a ground-truth variant of the EKF which uses IMU and Vicon measurements only. 
Besides the output of the EKF, additionally the estimates of VIO with and without predictions, all other input data to the EKF, and the Vicon ground-truth are recorded at 1 kHz to be able to assess the state estimate of other EKF variants on the runs. 
We compare variants and ablations of our approach including EKF with leg velocity measurements only \emph{ekf\_leg}, EKF with Vicon \emph{ekf\_vicon}, EKF with augmented VIO \emph{ekf\_vio+}, EKF with original VIO \emph{ekf\_vio}, augmented VIO and original VIO. 
For \emph{ekf\_leg} and \emph{ekf\_vio+} we tuned separate covariance parameters for the EKF empirically. For the variants \emph{ekf\_vicon} and \emph{ekf\_vio}, we use the same parameters like \emph{ekf\_vio+}.
We compute position error~(labeled as x, y, z) in meter, yaw~(labeled as yaw) and roll-pitch error~(labeled as gravity) in degree separately. 



\subsection{Trajectory Accuracy Evaluation for Indoor Experiments}
\subsubsection{Trotting Gait}

In the trotting gait, at least two feet of diagonal legs are always in contact with the ground.
The base link oscillates vertically with an amplitude of ca. 2\,cm. 
The RPE evaluation is summarized in Tab.~\ref{tab:result_trotting} and Fig.~\ref{fig:rpe_trotting}. 
The EKF with only leg velocity measurements (\emph{ekf\_leg}) shows significant drift in position and yaw orientation (avg. 0.333\,m x-y-pos., 3.902\,deg yaw at 50\,s).
Integrating predicted VIO measurements (\emph{ekf\_vio+}) reduces this drift strongly, reducing the horizontal position and the yaw error to avg. 0.039\,m and 0.552\,deg at 50\,s.
We also observe that upsampling the VIO with IMU predictions improves the accuracy of pure VIO.
Note that the data is further upsampled with the latest estimate to 1\,kHz for reference to show its performance as potential input to the controller.
For shorter time intervals below the gait cycle time (0.5\,s), fusing leg odometry in the EKF variants improves the accuracy of the pure VIO variants.
Fusing \emph{vio+} or ground truth with the leg odometry increases the roll pitch drift slightly towards \emph{ekf\_leg}, even though \emph{vio+} shows lower drift.
At larger time intervals, the EKF finds a trade-off with high accuracy in horizontal position and orientation.
VIO shows a small drift in height for trotting, which is also reflected by the RPE. 
Importantly, filtering leg kinematics and VIO allows for estimating the absolute height of the base with respect to the ground with high accuracy.

\subsubsection{Jumping Gait}


\begin{table}
	\caption{Jumping trajectory accuracy in RPE.}
	\label{tab:result_jumping}
	\centering
		\resizebox{\linewidth}{!}{
		\begin{tabular}{cccccccc}
			\toprule
			& & ekf\_leg & ekf\_vicon & ekf\_vio+ & ekf\_vio  & vio+ & vio\\
			\midrule
            \multirow{2}{*}{x and y\,(m)}& mean & 0.229 & 0.013 & 0.035 & 0.090 & 0.021 & 0.022\\
            & max & 1.286 & 0.054 & 0.175 & 0.358 & 0.124 & 0.123 \\
			\midrule
            \multirow{2}{*}{z\,(m)}& mean & 0.187 & 0.013 & 0.015 & 0.017 & 0.084 & 0.100\\
            & max & 0.806 & 0.086 & 0.107 & 0.131 & 0.610 & 0.646 \\
			\midrule
			\multirow{2}{*}{yaw\,(deg)}& mean & 4.923 & 0.277 & 0.632 & 1.728 & 0.365 & 0.540 \\
			& max & 32.928 & 2.848 & 3.754 & 9.765 & 2.262 & 3.236 \\
			\midrule
			\multirow{2}{*}{gravity\,(deg)}& mean & 0.686 & 0.900 & 0.923 & 0.860 & 0.711 & 1.646 \\
			& max & 3.357 & 3.360 & 3.368 & 3.212 & 3.769 & 7.845 \\
			\bottomrule
		\end{tabular}
		}
\end{table}

\begin{figure}[tb]
	\centering
	\includegraphics[width=0.99\linewidth]{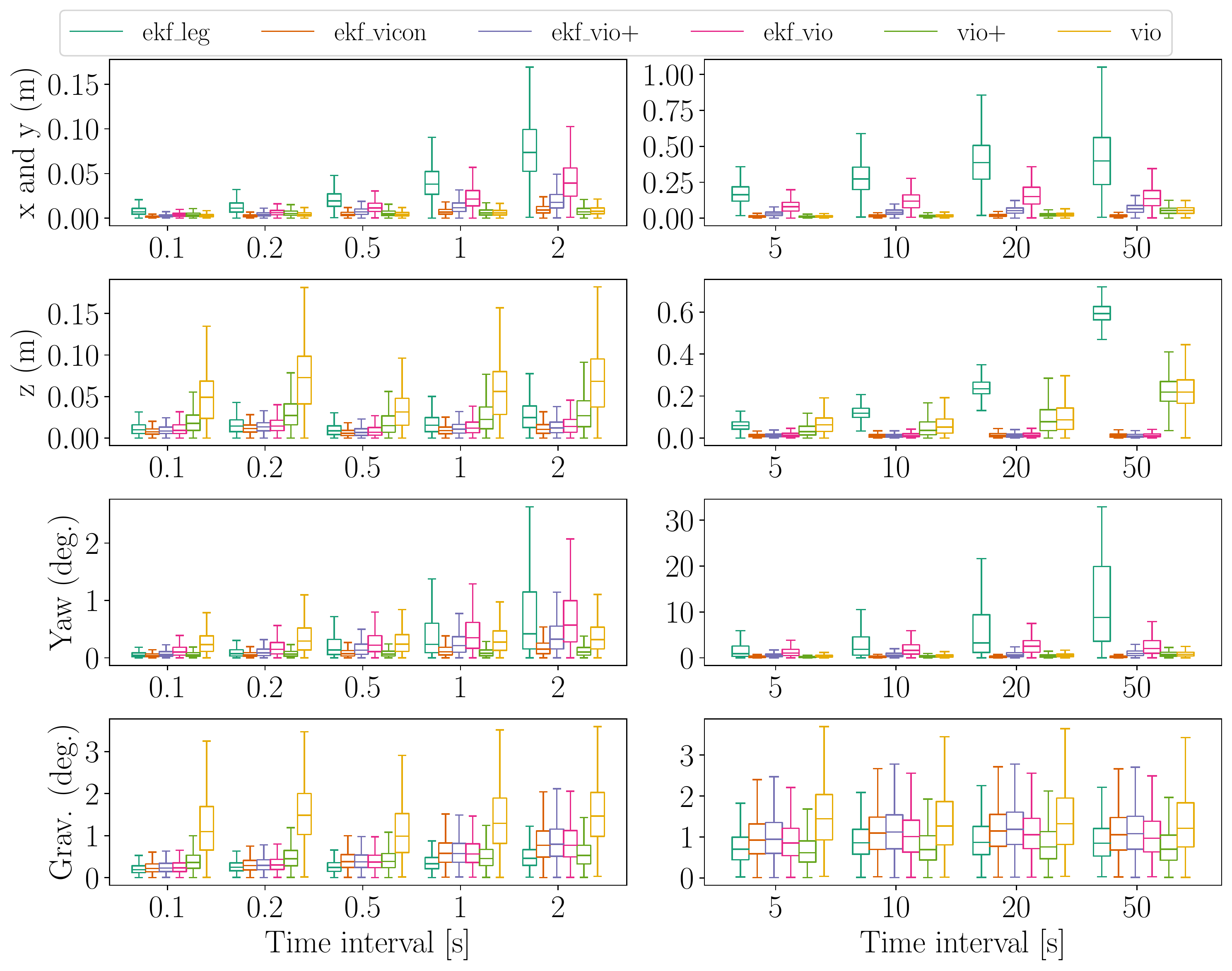}
	\caption{Jumping RPE for all time intervals.}
	\label{fig:rpe_jumping}
	\vspace{-0.3cm}
\end{figure}

In the jumping gait all four legs contact the ground at the same time during the landing and rebound phases. 
Each jump takes about 0.4\,s with a height of 12\,cm (robot base height change). 
In Table~\ref{tab:result_jumping} and Fig.~\ref{fig:rpe_jumping} we provide RPE results. 
It can be seen that despite the agile motion, our approach \emph{ekf\_vio+} can track the robot position and orientation.
The pure VIO shows significant drift in height due to the difficulty of tracking and reconstructing keypoints in the close vicinity of the robot and the larger noisy IMU accelerations.
This can be well compensated for by our EKF fusion approach (\emph{ekf\_vio+}, see also Fig.~\ref{fig:jump_height}).
The height bias estimate compensates the differences and enables control for the jumping gait. 
The bias takes about 1\,s to converge during the initialization phase in which the robot is standing before the jumping gait is started.
%
The yaw and horizontal position drift of \emph{ekf\_vio+} is slightly higher than in the trotting experiments.
It clearly improves over the drift of \emph{ekf\_leg}.

\begin{figure}[tb]
    \centering
    \includegraphics[width=0.99\linewidth]{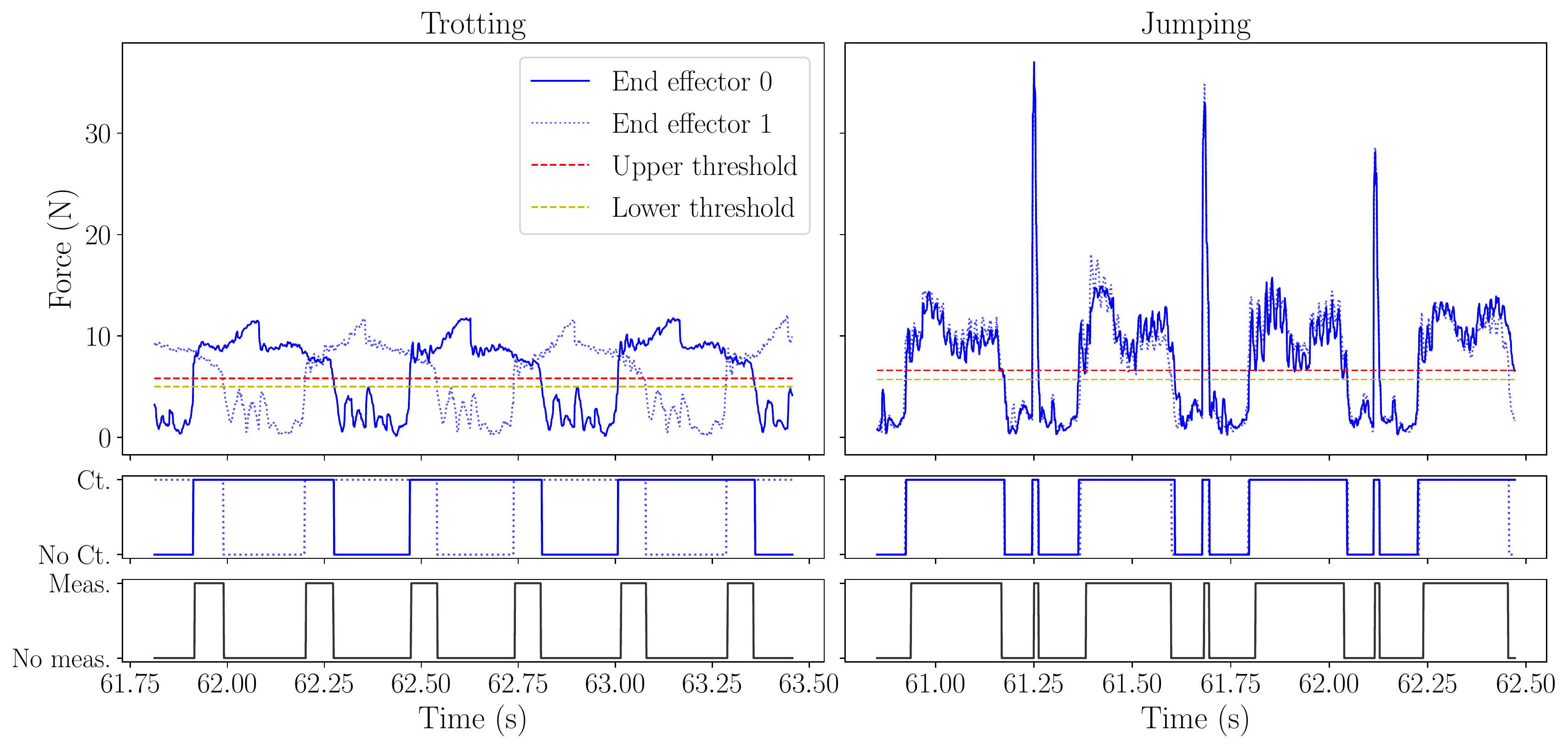}
    \caption{Contact detection for trotting and jumping gait for two end-effectors. The force estimate for jumping contains outliers that lead to false contact detection ($N_{\mathit{standing}}=3$).
    }
    \label{fig:contact}
    \vspace{-0.3cm}
\end{figure}

We also provide a qualitative assessment of the contact detection in Fig.~\ref{fig:contact}. 
For the jumping gait, high acceleration of the legs while pulling in the legs leads to high force estimates.
Our experiments demonstrate that the system can be sufficiently robust against these spurious false measurements for trotting and jumping at moderate speeds.
It is an interesting direction for future work to investigate more sophisticated ways of classifying contacts for dynamic gaits.
By setting higher contact duration thresholds ($N_{\mathit{contact}} = N_{\mathit{standing}} = 20$) for leg odometry and ground height measurements, the false contact detection can be avoided.
However, this also decreases the accuracy of the filter (mean rmse increases from 0.015\,m to 0.038\,m for z and from 0.632\,deg to 0.869\,deg for yaw), while \textit{ekf\_leg} fails.
 
\subsection{Outdoor Experiment}
We also tested our system outdoors on challenging asphalt and grass with slight slope for trotting and jumping gaits including gait switching with varying control speeds.
Please refer to the supplemental video
at \url{https://youtu.be/GFitG3TLEmw}
for footage from these runs.  
\section{CONCLUSIONS}
In this paper we present a lightweight EKF-based framework that fuses VIO estimate with leg odometry to calculate pose and velocity of the robot at high frequency. 
To compensate the delay and low rate from VIO we propose to use IMU  predictions to update the VIO state estimate such that the output of VIO is streamed at IMU rate with a significantly smaller delay and higher rate. 
Additionally, we compensate the drift of the height estimate by measuring height from leg kinematics and contact detection. 
We validate our approach with real-world experiments in both indoor and outdoor environments. 
The quantitative results of our experiments indicate that the low latency VIO with IMU prediction improves the accuracy of the EKF state estimate and the height measurement can prevent drift of the height estimate despite the existence of outliers in contact detection for the jumping gait. 
We also provide qualitative results for our system in challenging outdoor experiments. 
In these examples, our approach can estimate the robot state and perform trotting and jumping gaits including gait switching on different terrains. 
In future work, we aim to increase the robustness of our method and integrate terrain measurements to enable trajectory planning and control on complex terrain. 




%


\FloatBarrier
\bibliographystyle{IEEEtran}
\bibliography{literature}

\end{document}